%% file: root.tex
\documentclass[letterpaper, 10 pt, journal, twoside]{IEEEtran}

\IEEEoverridecommandlockouts                              

\usepackage{caption}
\usepackage{array}
\newcolumntype{C}[1]{>{\centering\arraybackslash}p{#1}}
\usepackage{siunitx}
\usepackage{bm} 
\usepackage{booktabs}
\usepackage{graphicx}
\usepackage{subcaption}
\usepackage{svg}      
\usepackage{amsmath}
\usepackage{amssymb}
\usepackage{bbm}
\usepackage{todonotes}
\usepackage{cite}
\usepackage{url}
\usepackage[hidelinks]{hyperref}
\usepackage{dsfont}

\DeclareMathOperator*{\argmin}{argmin}

\begin{document}

\title{SOCC-ICP: Semantics-Assisted Odometry based on Occupancy Grids and ICP}

\author{Johannes Scherer$^{1,2,^{*}}$, Sebastian Hirt$^{3}$, and Henri Mee{\ss}$^{1}$%
\thanks{Manuscript received: January 10, 2026; Revised April 5, 2026; Accepted May 7, 2026.}
\thanks{This paper was recommended for publication by Editor Ayoung Kim upon evaluation of the Associate Editor and Reviewers' comments.}
\thanks{$^{*}$Corresponding author ({\tt\footnotesize johannes.scherer.ivi@outlook.com}).}%
\thanks{$^{\text{1}}$Fraunhofer IVI, 85049 Ingolstadt, Germany.}%
\thanks{$^{\text{2}}$Technische Hochschule Ingolstadt, 85049 Ingolstadt, Germany.}%
\thanks{$^{\text{3}}$Ancud IT-Beratung GmbH, 90478 Nürnberg, Germany.}%
\thanks{Digital Object Identifier (DOI): see top of this page.}
}


\markboth{IEEE Robotics and Automation Letters. Preprint Version. Accepted May, 2026}
{Scherer \MakeLowercase{\textit{et al.}}: SOCC-ICP: Semantics-Assisted Odometry based on Occupancy Grids and ICP} 






\maketitle



\input{full_content}


\bibliographystyle{IEEEtran}   
\bibliography{library}         

\end{document}

%% file: full_content.tex
\begin{abstract}
Reliable pose estimation in previously unseen environments is a fundamental capability of autonomous systems. Existing LiDAR odometry methods typically employ point-, surfel-, or NDT-based map representations, which are distinct from the semantic occupancy grids commonly used for downstream tasks such as motion planning.
We introduce SOCC-ICP, a semantics-assisted odometry framework that jointly performs Semantic OCCupancy grid mapping and LiDAR scan alignment.
Each map voxel encodes geometric and semantic statistics, enabling adaptive point-to-point or point-to-plane ICP based on local planarity.
Further, the occupancy grid naturally filters dynamic objects through raycasting-based free-space updates.
Across diverse evaluation scenarios, SOCC-ICP achieves performance competitive with state-of-the-art LiDAR odometry and remains robust in geometrically degenerate environments, even in the absence of semantic cues.
When semantic labels are available, integrating them into map construction, downsampling, and correspondence weighting yields further accuracy gains.
By unifying odometry and semantic occupancy grid mapping within a single representation, SOCC-ICP eliminates redundant map structures and directly provides a map suitable for downstream robotic applications.
\end{abstract}



\begin{figure*}[!t]
    \centering
    \includegraphics[
      width=0.948\textwidth,
      trim=0 25mm 0 9mm,
      clip
    ]{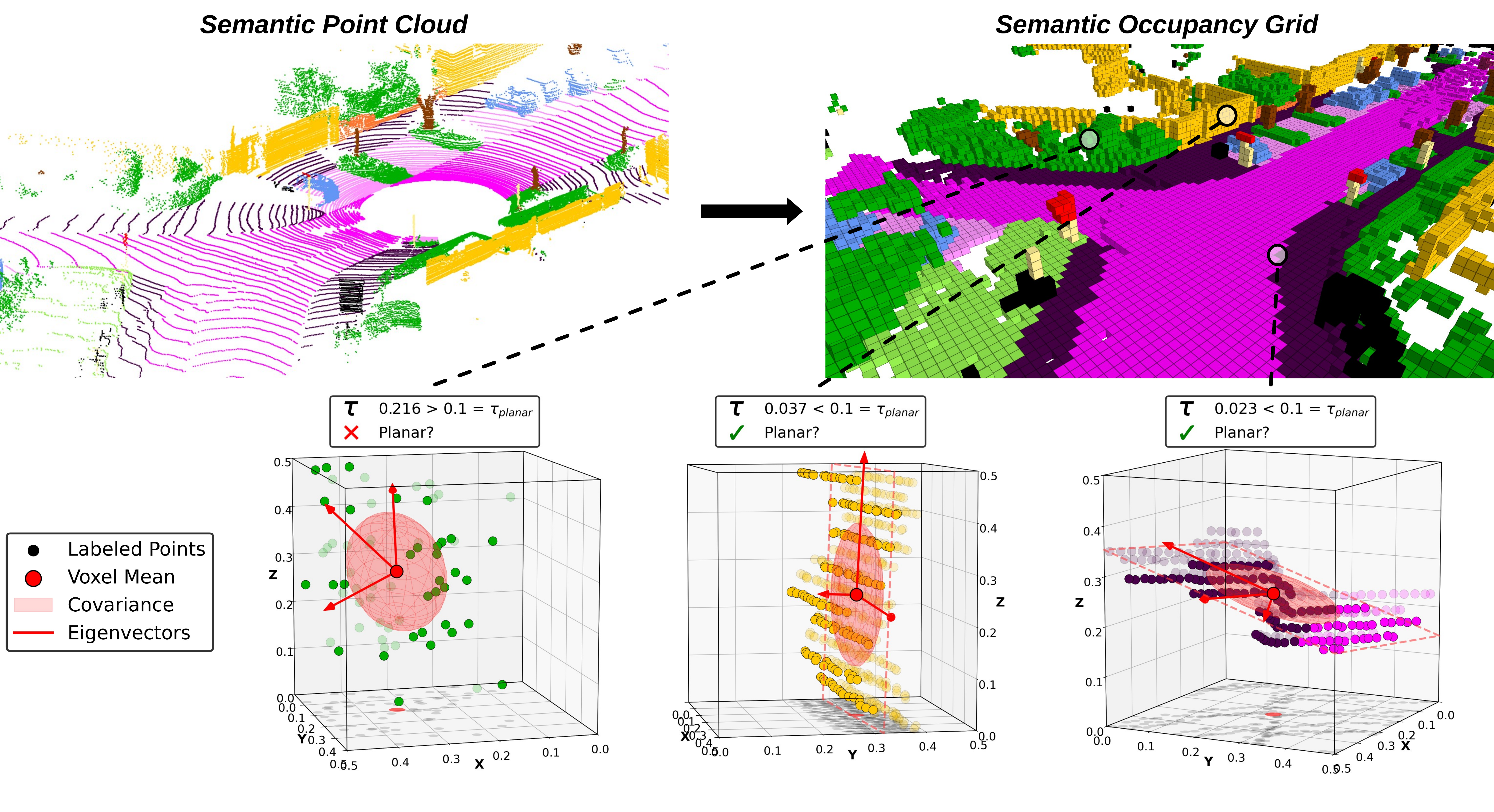}
    \caption{Overview of SOCC-ICP scan registration. A semantically segmented LiDAR point cloud is aligned with a semantic occupancy grid encoding voxel-level geometric and semantic statistics. Three representative voxels of size 0.5\,m are shown.}
    \label{fig:socc_icp}
\end{figure*}

\section{Introduction}
\IEEEPARstart{A}{ccurate} and robust odometry is a prerequisite for autonomous systems operating in complex and dynamic environments. Whether in autonomous driving~\cite{SA-LOAM}, aerial robotics~\cite{LOAMTree}, or GNSS-denied environments~\cite{GnssDeniedUAV}, robots rely on precise ego-motion estimation to support downstream tasks such as mapping and planning.
With the growing availability of LiDAR sensors, their robustness to illumination changes, and increasing range and point density, LiDAR-only odometry has received significant attention~\cite{surveyLidarOdometry}.
In particular, methods based on the Iterative Closest Point (ICP)~\cite{ICP} algorithm are widely used to estimate the transformation between successive LiDAR scans or against existing maps.
Recent systems such as KISS-ICP~\cite{KISS-ICP} and GenZ-ICP~\cite{GenZ-ICP} demonstrate that carefully designed ICP pipelines achieve state-of-the-art accuracy while remaining lightweight, computationally efficient, and minimally parameterized.

In parallel, occupancy grids are widely used as probabilistic representations of free and occupied space, supporting tasks such as path planning~\cite{occGridPathPlanning} and collision avoidance~\cite{OccGridAvoidance}, and, when enriched with semantic labels, enabling semantic reasoning~\cite{SemOCC2}.
While scan matching based on occupancy grid maps is well established in 2D~\cite{2DOCCSLAM1, OccupancySLAM}, 3D LiDAR odometry pipelines typically maintain separate map representations such as point clouds~\cite{KISS-ICP, GenZ-ICP}, surfel maps~\cite{SUMA}, (adaptive) voxel-based maps~\cite{VoxelMap}, or NDT distributions~\cite{NDT-LOAM}, leading to duplicated data structures and increased memory usage.
A 3D occupancy grid used for both mapping and odometry would simplify system architecture while maintaining coherent geometric and semantic information.
While prior work has used 3D occupancy grids for localization against pre-built maps~\cite{PreBuiltOcc} or as downstream mapping outputs after pose estimation~\cite{KISS-SLAM}, to the best of our knowledge, no existing LiDAR odometry system performs online scan-to-map registration on a 3D semantic occupancy grid.

We introduce SOCC-ICP, a semantics-assisted odometry framework that jointly performs semantic occupancy grid mapping and ICP-based scan alignment, demonstrating that a 3D occupancy grid is an effective map representation for LiDAR odometry.
SOCC-ICP maintains a rolling 3D occupancy grid map whose voxels encode geometric and semantic statistics, including sample mean, empirical covariance, occupancy probability, and a per-class semantic distribution. 
These voxel statistics enable ICP-based scan registration with adaptive selection between point-to-point and point-to-plane residuals based on local planarity (see Figure~\ref{fig:socc_icp}), while simultaneously constructing a semantic occupancy grid suitable for downstream tasks.
When available, LiDAR semantic segmentation~\cite{LSK3DNet} guides semantics-aware downsampling and correspondence weighting, while voxel occupancy probabilities and a robust kernel further refine the weighting scheme.
The occupancy grid also improves robustness by suppressing dynamic objects through raycasting-based free-space updates (see Figure~\ref{fig:cleaning_ray}).
Importantly, SOCC-ICP remains fully functional without semantic information.
Since the semantic occupancy grid is constructed as an integral part of pose estimation, SOCC-ICP directly provides the map representation required by downstream tasks such as planning and collision avoidance.

We evaluate SOCC-ICP on diverse real-world datasets spanning urban environments, highways, campus scenes, and indoor corridors to support three key claims: (i) even without semantic cues, SOCC-ICP achieves accuracy comparable to state-of-the-art LiDAR odometry methods while relying solely on an occupancy grid map representation, (ii) incorporating semantic information can yield additional accuracy gains, and (iii) the method maintains reliable performance in geometrically degenerate environments through adaptive selection of point-to-point and point-to-plane residuals. The full SOCC-ICP odometry system, including the integrated Radix library for semantic occupancy grid mapping, is available at \href{https://github.com/josch14/socc_icp}{\texttt{\small https://github.com/josch14/socc\_icp}}.

\section{Related Work}
\label{sec:related_work}

\subsection{LiDAR Odometry Estimation}
LiDAR odometry estimates robot motion by registering successive point cloud scans. Existing approaches generally fall into two categories. Scan-to-scan methods align consecutive scans directly, while scan-to-map methods register each incoming scan against an accumulated map representation.
While scan-to-scan matching is inherently limited in accuracy, scan-to-map methods generally yield lower drift and represent the dominant paradigm in modern systems~\cite{surveyLidarOdometry}.
Among scan-to-map approaches, direct methods typically rely on ICP variants that minimize geometric residuals between the incoming scan and a reference map.
Prominent examples include KISS-ICP~\cite{KISS-ICP}, which employs point-to-point ICP with adaptive correspondence thresholds and robust outlier handling, and GenZ-ICP~\cite{GenZ-ICP}, which addresses performance degradation in geometrically degenerate environments by adaptively weighting point-to-point and point-to-plane residuals.
Alternative formulations include NDT-based methods~\cite{NDT}, which formulate registration as likelihood maximization over probabilistic voxel distributions, and continuous-time extensions such as CT-ICP~\cite{CT-ICP}, which accounts for intra-scan motion distortion via continuous-time trajectory modeling.
Feature-based methods, by comparison, perform registration on reduced representations derived from geometrically informative primitives. For instance, LOAM~\cite{LOAM} extracts edge and planar feature points, SuMa~\cite{SUMA} leverages surface normals computed from a surfel-based map, and MULLS~\cite{MULLS} jointly registers multi-class feature points such as ground, facade, and pillar primitives.
SOCC-ICP adopts efficient and robust direct ICP formulations, but replaces the conventional point cloud map with a 3D occupancy grid that unifies scan registration and mapping within a single probabilistic data structure.

\begin{figure*}[!htbp]
    \centering
    \includegraphics[
      width=1.00\textwidth,
      trim=25 0 0 0mm,
      clip
    ]{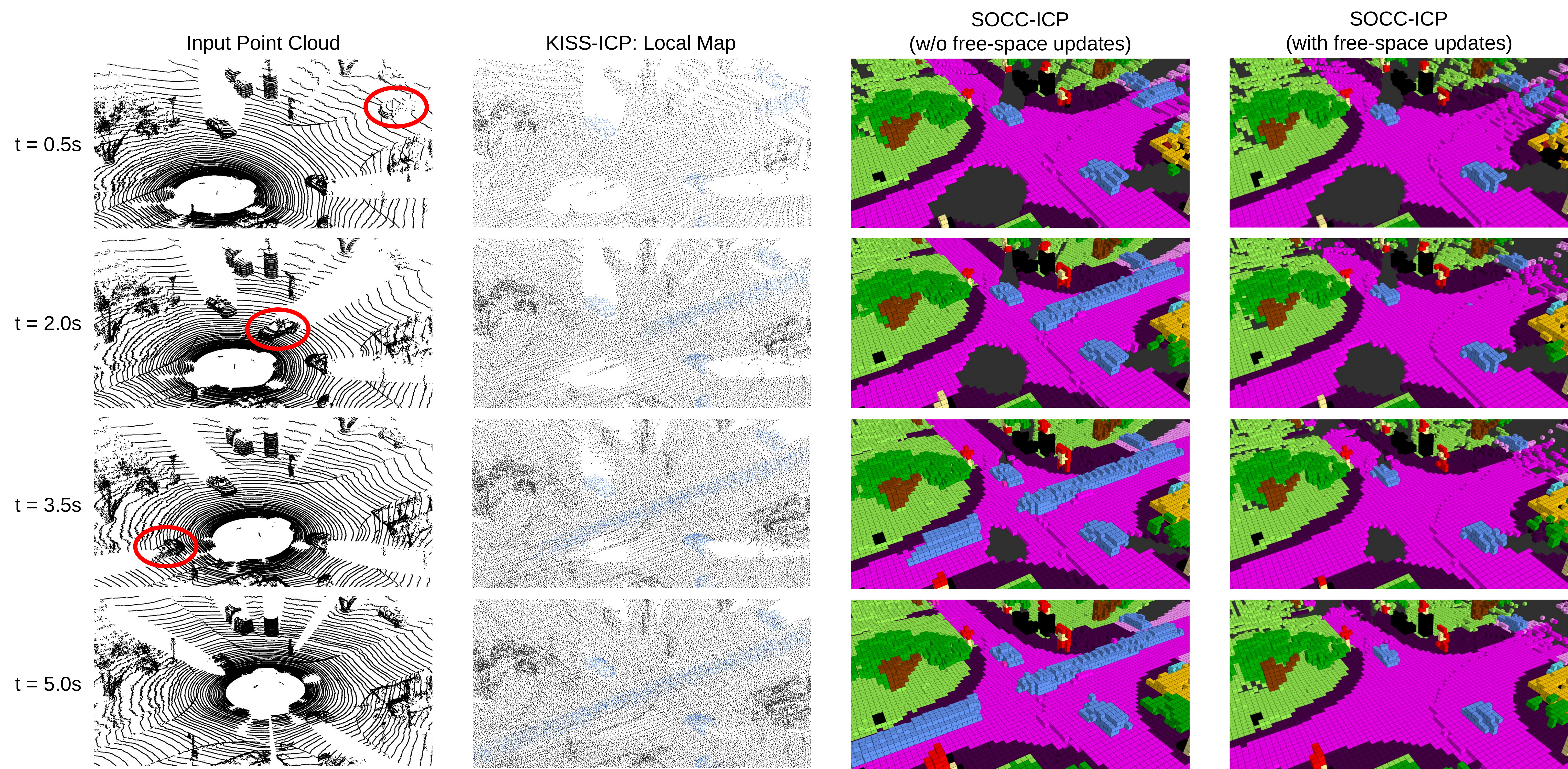}
    \caption{Local maps in approaches such as KISS-ICP retain persistent ghosted traces, which can degrade odometry accuracy (second column). In contrast, SOCC-ICP handles dynamic objects by automatically removing the corresponding voxel information from the 3D occupancy grid via raycasting-based free-space updates (fourth column).
    }
    \label{fig:cleaning_ray}
\end{figure*}

\subsection{Map Representations in LiDAR Odometry}
The choice of map representation is a key design decision in LiDAR odometry. Common representations include downsampled point clouds stored in voxel grids~\cite{KISS-ICP, GenZ-ICP}, surfel maps~\cite{SUMA, SUMA++}, sparse feature maps composed of edge and planar primitives~\cite{LOAM, SA-LOAM}, and Gaussian voxel distributions as used in NDT~\cite{NDT, NDT-LOAM}. VoxelMap~\cite{VoxelMap} adopts an adaptive voxel-based representation that encodes planar structures within a probabilistic map.

The use of occupancy grids for LiDAR odometry has been explored primarily in 2D. \cite{2DOCCSLAM1} projects incoming 3D point clouds onto the ground plane, constructs 2D occupancy maps, and registers consecutive maps in a scan-to-scan fashion via phase correlation. In contrast, \cite{OccupancySLAM} performs scan-to-map registration against a 2D occupancy grid, using 2D laser scans in an offline setting. 
In the 3D domain, the use of occupancy grid maps as the registration target remains largely unexplored. Existing approaches either localize against pre-built 3D occupancy maps without incremental scan alignment~\cite{PreBuiltOcc}, or employ occupancy grids solely as a downstream mapping output after pose estimation~\cite{KISS-SLAM}. NDT-based methods such as NDT-LOAM~\cite{NDT-LOAM} represent local geometry using Gaussian voxel models but do not maintain a probabilistic occupancy grid, and therefore lack an explicit representation of free space.

In contrast, SOCC-ICP uses a 3D occupancy grid as a unified map representation for both ICP-based scan registration and mapping. Rather than formulating registration over occupancy values directly, geometric primitives, specifically point information and surface normals, are extracted from the voxel statistics for standard ICP alignment, while the occupancy grid itself provides additional capabilities such as dynamic object removal and occupancy-weighted correspondences. To the best of our knowledge, SOCC-ICP is the first LiDAR odometry framework to perform online scan-to-map registration on a 3D semantic occupancy grid.

\subsection{Semantics-Assisted Odometry Estimation}
Several LiDAR odometry and SLAM systems incorporate semantic information to improve robustness in dynamic or ambiguous environments. 
SuMa++~\cite{SUMA++} integrates semantic labels into a surfel-based mapping and registration pipeline to suppress points associated with dynamic object classes. 
SA-LOAM~\cite{SA-LOAM} augments LOAM with semantics-assisted matching, class-specific downsampling, and semantic planar constraints. 
SAGE-ICP~\cite{SAGE-ICP} employs point-wise semantic labels to guide subsampling, remove dynamic vehicles, weight correspondences based on semantic similarity, and maintain a semantics-aware local map.
These approaches share a common strategy of identifying and suppressing potentially dynamic classes based on semantic labels. However, semantic class membership alone does not reliably indicate motion. A parked vehicle, for instance, is semantically indistinguishable from a moving one, and class-level filtering can therefore degrade odometry accuracy~\cite{SUMA++}. Instead, SOCC-ICP handles dynamic behavior implicitly through free-space updates on the occupancy grid, while semantic cues, when available, further refine downsampling and registration.

\section{Radix: A Semantic Occupancy Grid Mapping Library}
\label{method:occ_grid_mapping}

Bonxai~\cite{Bonxai} is an open-source C++ library for large-scale volumetric mapping based on a sparse 3D voxel grid, designed as a high-performance alternative to octree-based probabilistic frameworks such as OctoMap~\cite{Octomap}. 
Inspired by OpenVDB~\cite{OpenVDB}, it stores only active voxel regions in a compact three-level hierarchy, enabling fast, near-constant-time access and updates.
Bonxai natively supports probabilistic occupancy mapping, including point insertion, raycasting, and voxel traversal.

We introduce Radix, an open-source library that extends Bonxai's sparse voxel grid into a configurable semantic occupancy mapping framework, integrated as a ROS~2 node. 
In this work, we leverage Radix to maintain the map representation underlying our odometry pipeline.
Radix incrementally fuses pose-synchronized point clouds, optionally enriched with semantic labels, into the occupancy grid.
A key feature of Radix is its behavior-based design, allowing custom mapping logic through minimal, well-scoped modifications. 
For SOCC-ICP, this includes a voxel data structure that, in addition to occupancy probabilities, stores geometric statistics and semantic class distributions.
Occupancy is updated via the inverse sensor model with raycasting-based free-space updates.
Radix also exposes ROS~2 services such as spatial chunk queries that enable SOCC-ICP to efficiently retrieve local map regions for scan registration, and semantic image rendering.

In the following, we detail the semantic occupancy grid mapping behavior implemented within Radix for SOCC-ICP, which aggregates geometric and semantic information per voxel to provide the scan registration pipeline (see Sec.~\ref{method:socc_icp}) with the representations required for accurate and robust odometry estimation.
We denote the occupancy grid map as $\mathcal{V} = \{\, \mathcal{V}_i \mid i \in \mathcal{I} \,\}$, where each voxel stores a tuple
\begin{equation}
\mathcal{V}_i =
\bigl(
\boldsymbol{\mu}_i,\;
\boldsymbol{\Sigma}_i,\;
\mathbf{q}^{(1)}_i,\;
n_i,\;
p^{\text{occ}}_i,\;
\mathbf{p}_i,\;
c_i
\bigr).
\end{equation}
Here, 
$\boldsymbol{\mu}_i \in \mathbb{R}^3$ and 
$\boldsymbol{\Sigma}_i \in \mathbb{R}^{3\times 3}$ denote the sample mean and empirical covariance of all points inserted into voxel $i$, which summarize its local surface geometry and are updated incrementally after each frame.
The vector $\mathbf{q}^{(1)}_i \in \mathbb{R}^3$ stores the first point inserted into the voxel and is retained as a fixed reference for scan alignment, while
$n_i \in \mathbb{N}$ denotes the number of accumulated voxel points. 
The scalar $p^{\text{occ}}_i \in [0,1]$ represents the occupancy probability.
Semantic information is captured by the per-class probability vector $\mathbf{p}_i = (p_{i,0}, \dots, p_{i,K}) \in [0,1]^{K+1}$, with $\sum_{k=0}^{K} p_{i,k} = 1$, and the corresponding semantic label $c_i = \arg\max_{k}\, p_{i,k}$.

Voxel occupancy probabilities are updated via the Bayesian log-odds formulation, as adopted in OctoMap~\cite{Octomap}. 
Each voxel maintains a log-odds value $L_i = \ell(p_i^{\text{occ}})$, where $\ell(p) = \log\!\left(\frac{p}{1 - p}\right)$. With a neutral prior of $p_i^{\text{occ}} = 0.5$, the update reduces to adding class-dependent increments: hits increase $L_i$ by $\ell(p^{\text{hit}}_{c_i})$, while raycasting-based free-space observations decrease $L_i$ by $\ell(p^{\text{miss}}_{c_i})$. Class-dependent miss probabilities enable faster decay of potentially dynamic classes (e.g., \textit{car}) while preserving evidence for static structures (e.g., \textit{building}).

As introduced, each voxel $\mathcal{V}_i$ maintains a semantic class distribution \(\mathbf{p}_i = (p_{i,0}, \dots, p_{i,K})\) over $K{+}1$ semantic classes. Given an incoming LiDAR scan, we collect the subset of hit points whose coordinates fall within $\mathcal{V}_i$. For the $N_i$ points in this set, we count the occurrences of each semantic label and compute their empirical frequencies
\begin{equation}
\hat{p}_{i,k} = \frac{1}{N_i} \sum_{j=1}^{N_i} \mathds{1}[c_j = k],
\end{equation}
where $c_j$ denotes the semantic label of the $j$-th point and $\mathds{1}[\cdot]$ is the indicator function. The voxel's semantic distribution is then updated using an exponential moving average (EMA),
\begin{equation}
p_{i,k}' = \alpha\, p_{i,k} + (1 - \alpha)\, \hat{p}_{i,k},
\end{equation}
where $\alpha \in [0, 1]$ retains past evidence and $1{-}\alpha$ weights the newly observed label statistics.
The voxel's semantic label is updated to the maximum-probability class. 
Unlike Bayesian evidence accumulation (e.g., \cite{SemOCC1, SemOCC2}), the EMA bounds effective memory and enables rapid adaptation when a voxel's dominant class changes. While not strictly Bayesian optimal, the approach is numerically stable and well suited for real-time semantic mapping.

\section{SOCC-ICP: Semantics-Assisted Odometry Based on Occupancy Grids and ICP}
\label{method:socc_icp}

We estimate the trajectory of a moving LiDAR sensor by incrementally registering successive scans. Given an input point cloud $\mathcal{Q} = \{\mathbf{q}_j \mid \mathbf{q}_j \in \mathbb{R}^3\}$ recorded at time $t$, each scan is processed by a LiDAR semantic segmentation model to obtain a semantic point cloud $\mathcal{S} = \bigl\{ (\mathbf{q}_j, c_j) \mid \mathbf{q}_j \in \mathbb{R}^3,\; c_j \in \{0,\dots,K\}\bigr\}$, where $K{+}1$ is the number of semantic classes and 0 corresponds to the \textit{unlabeled} class. The goal is to estimate the sensor pose $\mathbf{T} = (\mathbf{R}, \mathbf{t}) \in SE(3)$ that best aligns $\mathcal{S}$ with the semantic occupancy grid $\mathcal{V}$ (see Sec.~\ref{method:occ_grid_mapping}) built up to the current time.

SOCC-ICP builds on the scan alignment approach of KISS-ICP~\cite{KISS-ICP}, while extending it with several key components.
A constant-velocity motion model is used to predict the sensor motion and deskew the incoming scan based on per-point relative timestamps. 
The segmented point cloud is then downsampled using a semantics-aware strategy (Sec.~\ref{subsec:sem_pc_downsampling}).
Correspondences are subsequently established between the downsampled semantic point cloud and the semantic occupancy grid map.
Point-to-point and point-to-plane residuals are adaptively combined based on local surface planarity, and further weighted using a robust kernel, occupancy probability, and semantic consistency (Sec.~\ref{subsec:corr_search_and_weighting}).
Finally, the occupancy grid map is updated (see Sec.~\ref{method:occ_grid_mapping}) using the full semantic point cloud transformed by the estimated pose.
In the absence of semantic predictions, all points are assigned the \textit{unlabeled} class and SOCC-ICP operates as a purely geometric method.

\subsection{Semantic-Aware Point Cloud Downsampling}
\label{subsec:sem_pc_downsampling}

To obtain a compact yet geometrically informative point cloud for registration, we downsample the incoming semantic point cloud using a semantics-aware voxel downsampling strategy similar to SAGE-ICP~\cite{SAGE-ICP}, adapting the voxel size based on each input point's semantic class.
Rather than suppressing dynamic objects (handled by free-space updates, see Figure~\ref{fig:cleaning_ray}), the objective is to preserve fine geometric detail for discriminative classes while reducing the influence of regions that contribute little to robust scan matching.

We perform voxel downsampling on each class-specific subset $\mathcal{S}_c = \{\, (\mathbf{q}_j, c_j) \in \mathcal{S} \mid c_j = c \,\}$ of the input semantic point cloud $\mathcal{S}$, using an effective voxel size $v_c = \alpha_c v_\text{adapt}$, where $v_\text{adapt}$ is the current adaptive voxel size (see~\cite{KISS-ICP} for details) and $\alpha_c$ is a class-dependent downsampling factor that defaults to $1.0$.
Unlike KISS-ICP, we found it beneficial to retain the mean input point of each voxel rather than a single representative input point, yielding a downsampling operation instead of subsampling. The final downsampled semantic point cloud $\mathcal{S}^*$ is obtained by merging the class-wise results.

The downsampling factors $\alpha_c$ are selected to preserve high-resolution static geometry while maintaining balanced sparsity. Human-related classes such as \textit{person}, \textit{bicyclist}, and \textit{motorcyclist} are removed entirely ($\alpha_c = 0.0$), as they rarely provide stable geometric constraints. Small but discriminative structures such as \textit{poles} and \textit{traffic-signs} are retained at higher resolution ($\alpha_c = 0.75$). 
Finally, we found it beneficial to retain a higher sampling density for ground classes such as \textit{road}, \textit{parking}, \textit{sidewalk}, and \textit{terrain} ($\alpha_c = 0.8$). We attribute this to the characteristic LiDAR scanning pattern in autonomous driving scenarios, where ground returns are often comparatively sparse relative to vertical structures. These values can be adapted to other robotic platforms and environments.

\subsection{Semantic-Assisted Data Association}
\label{subsec:corr_search_and_weighting} 
We register the downsampled semantic point cloud $\mathcal{S}^*$ to the semantic occupancy grid map built up so far, denoted by $\mathcal{V} = \{\, \mathcal{V}_i \mid i \in \mathcal{I}_{\text{occ}} \,\}$, where $\mathcal{I}_{\text{occ}} \subset \mathcal{I}$ denotes the set of occupied voxel indices ($p^{\text{occ}}_i \ge 0.5$). As described in Section~\ref{method:occ_grid_mapping}, each voxel $\mathcal{V}_i$ stores the tuple of statistics $\bigl(\boldsymbol{\mu}_i,\; \boldsymbol{\Sigma}_i,\; \mathbf{q}^{(1)}_i,\; n_i,\; p^{\text{occ}}_i,\; \mathbf{p}_i,\; c_i \bigr)$.
During each ICP iteration, we compute correspondences between the downsampled cloud $\mathcal{S}^*$ and the set of first-inserted points of occupied voxels, $\{\mathbf{q}^{(1)}_i  \mid i \in \mathcal{I}_{\text{occ}}\}$, using a nearest-neighbor search. 
We deliberately avoid voxel centroids as references, as they shift with the integration of new measurements. Instead, fixed first-inserted points provide a stable geometric scaffold for ICP, preserving local surface structure within each voxel and yielding more consistent scan registration in practice.
Only correspondences whose Euclidean distance falls below an adaptive threshold are retained, with the threshold dynamically adjusted based on the predicted motion and current odometry uncertainty~\cite{KISS-ICP}.

For each of the $N$ remaining correspondences, we compute a residual $\mathbf{e}_i$. The goal is to find the pose $\mathbf{T} \in SE(3)$ aligning the scan $\mathcal{S}$ with the occupancy grid map $\mathcal{V}$, estimated iteratively through relative increments $\Delta\hat{\mathbf{T}}$ that each minimize
\begin{equation}
\Delta\mathbf{\hat{T}} = \argmin_{\Delta\mathbf{T} \in SE(3)}
\mathcal{J}(\Delta\mathbf{T}),
\end{equation}
where the cost function $\mathcal{J}$ accumulates the squared residuals over all correspondences,
\begin{equation}
\mathcal{J}(\Delta\mathbf{T}) = 
\sum_{i=1}^N \left\|\mathbf{e}_i\right\|_2^2.
\end{equation}

To overcome the limitations of relying on a single error metric, we follow GenZ-ICP~\cite{GenZ-ICP} and adaptively combine point-to-point and point-to-plane residuals to leverage their complementary strengths. Correspondences between semantic points $\mathbf{s} \in \mathcal{S}^*$ and voxel reference points $\mathbf{q}^{(1)}$ are classified as \textit{planar} or \textit{non-planar}. For voxels containing at least five points, we compute the eigenvalues $\lambda_1 > \lambda_2 > \lambda_3$ and eigenvectors of the voxel covariance matrix $\boldsymbol{\Sigma}$. We then evaluate the local surface variation~\cite{LSV} defined as $\tau = \frac{\lambda_3}{\lambda_1 + \lambda_2 + \lambda_3}$,
where lower values indicate flatter and more consistent surface structure. A correspondence is classified as planar if the associated voxel exhibits $\tau < \tau_{\text{planar}}$.

A planar correspondence indicates that the local 3D structure represented by the voxel is sufficiently planar. In this case, we compute a point-to-plane residual $\mathbf{e}_\text{pl}$ using the semantic point $\mathbf{s} \in \mathcal{S}^*$, the corresponding voxel reference point $\mathbf{q}^{(1)}$, and the surface normal given by the eigenvector associated with the smallest eigenvalue of $\boldsymbol{\Sigma}$. Otherwise, the voxel is classified as non-planar, and a point-to-point residual $\mathbf{e}_\text{po}$ is computed between $\mathbf{s}$ and $\mathbf{q}^{(1)}$. A detailed derivation of the Jacobian-based linearization for both residuals is provided in~\cite{GenZ-ICP}. The residuals are then combined in the cost function
\begin{equation}
\mathcal{J}(\Delta\mathbf{T}) =
\alpha 
\sum_{i=1}^{N_{\text{pl}}} \left\|\mathbf{e}_{\text{pl},i}\right\|_2^2
+ (1 - \alpha)
\sum_{i=1}^{N_{\text{po}}} \left\|\mathbf{e}_{\text{po},i}\right\|_2^2 ,
\end{equation}
where $\alpha = \frac{N_{\text{pl}}}{N_{\text{pl}} + N_{\text{po}}} \in [0,1]$ is an adaptive weight determined by the number of planar ($N_{\text{pl}}$) and non-planar ($N_{\text{po}}$) correspondences. This formulation enables automatic adjustment to changing scene geometry, smoothly transitioning between structured and unstructured environments and improving robustness to geometric variability~\cite{GenZ-ICP}.

Each residual is further weighted using three complementary factors.
(i) We apply the Geman--McClure robust kernel $w_{\text{gm}} = \rho_{\text{gm}}(\mathbf{e}) \in [0,1]$, which suppresses outliers while preserving the influence of well-aligned correspondences~\cite{KISS-ICP}.
(ii) Voxel-level geometric confidence is incorporated through the occupancy probability $p^{\text{occ}} \in [0,1]$, yielding the weight $w_{\text{occ}} = (p^{\text{occ}})^{\gamma} \in [0,1]$, where $\gamma \ge 0$ controls decay. Note that correspondence search is restricted to voxels with $p^{\text{occ}} \ge 0.5$.
(iii) Semantic consistency is enforced between a semantic input point $\mathbf{s} = (\mathbf{q},c) \in \mathcal{S}^*$ and its associated map voxel $\mathcal{V}_{i}$ with semantic label~$c_{i}$. A correspondence is considered a semantic match if the labels agree or if either is \textit{unlabeled}:
\begin{equation}
\mathds{1}_{\text{match}}(c,c_{i}) =
\begin{cases}
1, & c = c_{i} \;\; \text{or} \;\; c=0 \;\; \text{or} \;\; c_{i}=0, \\[4pt]
0, & \text{otherwise}.
\end{cases}
\end{equation}
Let $p_{i}$ denote the probability of the most likely semantic class of voxel $\mathcal{V}_{i}$. The semantic weight is then defined as 
\begin{equation}
w_{\text{sem}} =
w_{\text{lower}}
+ \mathds{1}_{\text{match}}(c,c_{i})\,(1 - w_{\text{lower}})\, p_{i}  \in [0,1],
\end{equation}
where $w_{\text{lower}} \in [0,1]$ is a fixed weight for non-matching labels. The final correspondence weight is given by
\begin{equation}
w = w_{\text{gm}} \cdot w_{\text{occ}} \cdot w_{\text{sem}} \in [0,1],
\end{equation}
which jointly emphasizes stable, consistently observed, and semantically reliable regions while down-weighting uncertain or transient voxels. The cost function finally becomes
\begin{equation}
\mathcal{J}(\Delta\mathbf{T}) = 
\alpha 
\sum_{i=1}^{N_{\text{pl}}} w_i \left\|\mathbf{e}_{\text{pl},i}\right\|_2^2
+ (1 - \alpha)
\sum_{i=1}^{N_{\text{po}}} w_i \left\|\mathbf{e}_{\text{po},i}\right\|_2^2 .
\end{equation}
This objective is minimized at each ICP iteration to obtain the pose update $\Delta\hat{\mathbf{T}}$, and the cycle of downsampling, correspondence search, and optimization is repeated until convergence.  


\section{Experimental Evaluation}
\label{sec:exp}
We design the experimental evaluation to validate our three key claims that SOCC-ICP (i) achieves accuracy comparable to state-of-the-art LiDAR odometry methods in purely geometric settings across diverse scenarios, (ii) attains additional accuracy gains when semantic information is incorporated, and (iii) remains robust in degenerate environments.

\subsection{Experimental Setup}
We first evaluate SOCC-ICP on the KITTI Odometry Benchmark~\cite{KITTI}, which comprises 11 sequences covering structured urban environments and highways. The evaluation serves two purposes: to compare SOCC-ICP against state-of-the-art LiDAR odometry methods, with and without semantic assistance, and to assess the impact of incorporating semantic information into odometry estimation. On KITTI, SOCC-ICP uses semantic labels predicted by LSK3DNet~\cite{LSK3DNet}, which achieves 75.6\% mIoU on the SemanticKITTI~\cite{SemanticKITTI} test set across 19 classes in the single-scan track.

KITTI is the only dataset on which we evaluate SOCC-ICP with semantic assistance, as it provides both ground-truth odometry and semantic labels for training LiDAR segmentation models.
To demonstrate that SOCC-ICP also performs reliably without semantic cues, we further evaluate on the MulRan dataset~\cite{mulran}, recorded with a different LiDAR sensor in similar autonomous-driving scenarios, and on the Newer College dataset~\cite{NewerCollege}, which features handheld LiDAR scans in campus environments. 
For geometrically degenerate settings, we follow GenZ-ICP and use the Corridor1 and Corridor2 sequences from the Ground-Challenge dataset~\cite{GroundChallenge}, as well as the Long Corridor sequence from the SubT-MRS dataset~\cite{SubT-MRS}. 

For the KITTI, MulRan, and Newer College benchmarks, we report results from the respective original publications, with a few exceptions. On KITTI (Table~\ref{tab:results_kitti}), we reproduced the KISS-ICP results, while per-sequence metrics for GenZ-ICP were provided by the authors.\footnote{Per-sequence metrics for GenZ-ICP on KITTI: \url{https://github.com/cocel-postech/genz-icp/issues/14}.}
Results for MULLS on MulRan and Newer College, as well as CT-ICP on Newer College, are taken from~\cite{KISS-ICP}. For the Ground-Challenge and SubT-MRS corridor sequences, we use the results reported in~\cite{GenZ-ICP}, where all methods were fine-tuned for best performance. We also report results for complete SLAM systems such as CT-ICP, which include loop closure. In contrast, similar to KISS-ICP and GenZ-ICP, SOCC-ICP performs pure scan registration without loop closure.

On the KITTI Odometry Benchmark, we follow the standard evaluation protocol and report the relative translational error (RTE) and relative rotational error (RRE), averaged over subsequences of 100--800\,m. 
For the MulRan and Newer College datasets, we compute RTE using the same procedure. In the Ground-Challenge and SubT-MRS degenerate environments, we report both absolute pose error (APE) and relative pose error (RPE) for the translational component over the full sequences using the EVO evaluator~\cite{evoPackage}.

\subsection{Implementation}
SOCC-ICP is implemented on ROS~2 Humble, with Radix handling semantic occupancy grid mapping in a separate node and scan registration currently performed in Python.
The current system is a proof of concept and not optimized for efficiency. 
Despite the additional computational overhead introduced by semantic occupancy grid mapping, real-time performance remains achievable with a native implementation.

\subsection{Hyperparameters}
The parameters of SOCC-ICP are selected for autonomous driving scenarios and verified on the KITTI dataset. For the occupancy grid, we use a voxel size of 0.5\,m, an EMA decay rate of $\alpha = 0.8$, and default hit and miss probabilities of $p^\text{hit} = 0.55$ and $p^\text{miss} = 0.49$. When semantic information is available, class-specific miss probabilities of $p^\text{miss}_{\text{static}} = 0.498$ and $p^\text{miss}_{\text{moving}} = 0.475$ are applied to static (e.g., \textit{building}) and potentially moving (e.g., \textit{car}) classes, respectively. Within registration, for semantics-aware voxel downsampling we set $\alpha_c = 0.75$ for \textit{poles} and \textit{traffic-signs}, $0.8$ for ground classes, $0.0$ for human classes, and $1.0$ otherwise. We further use a planarity threshold of $\tau_{\text{planar}}=0.1$. Within correspondence weighting, we set $\gamma = 1.5$ and $w_{\text{lower}} = 0.25$. All parameters required for SOCC-ICP without semantic assistance are kept identical across datasets, except that on MulRan we use $p^\text{miss} = 0.475$, and on the Ground-Challenge and SubT-MRS sequences we use a voxel size of 0.2\,m and $p^\text{miss} = 0.485$.

\subsection{Performance on the KITTI Odometry Benchmark}

\begin{table*}[!t]
\centering
\caption{Quantitative results on the KITTI Odometry Benchmark~\cite{KITTI}. We report RTE in \% and RRE in deg/100m. Methods marked with $^\ast$ are semantics-assisted. SLAM systems and sequences with loop closures are marked with °. The best and second best performing odometry methods (excluding SLAM systems) are highlighted in \textbf{bold} and \underline{underline}, respectively.}
\label{tab:results_kitti}

{\scriptsize 
\setlength{\tabcolsep}{3.8pt} 
\renewcommand{\arraystretch}{1.0} 
\begin{tabular}{c|ccccccccccc|c}
\toprule
 & & &  &  &  & \textbf{Sequence} &  &  &  &  &  & \\ 
 & \textbf{00°} & \textbf{01} & \textbf{02°} & \textbf{03} & \textbf{04} & \textbf{05°} & \textbf{06°} & \textbf{07°} & \textbf{08°} & \textbf{09°} & \textbf{10} & \textbf{Avg.} \\ 
\textbf{Method} & \textbf{urban} & \textbf{highway} & \textbf{urban} & \textbf{country} & \textbf{country} & \textbf{country} & \textbf{urban} & \textbf{urban} & \textbf{urban} & \textbf{urban} & \textbf{country} & \textbf{RTE$\mathbf{/}$RRE} \\ 
\midrule
MULLS°~\cite{MULLS} & 0.54$/$0.13 & 0.62$/$0.09 & 0.69$/$0.13 & 0.61$/$0.22 & 0.35$/$0.08 & 0.29$/$0.07 & 0.29$/$0.08 & 0.27$/$0.11 & 0.83$/$0.17 & 0.51$/$0.12 & 0.61$/$0.19 & 0.52$/$0.13 \\ 
CT-ICP°~\cite{CT-ICP} & 0.49$/$\phantom{0.00} & 0.76$/$\phantom{0.00} & 0.52$/$\phantom{0.00} & 0.72$/$\phantom{0.00} & 0.39$/$\phantom{0.00} & 0.25$/$\phantom{0.00} & 0.27$/$\phantom{0.00} & 0.31$/$\phantom{0.00} & 0.81$/$\phantom{0.00} & 0.49$/$\phantom{0.00} & 0.48$/$\phantom{0.00} & 0.53$/$\phantom{0.00}\\ 
SuMa°~\cite{SUMA} & 0.68$/$0.23 & 1.70$/$0.54 & 1.20$/$0.48 & 0.74$/$0.50 & 0.44$/$0.27 & 0.43$/$0.20 & 0.54$/$0.30 & 0.74$/$0.54 & 1.20$/$0.38 & 0.62$/$0.22 & 0.72$/$0.32 & 0.83$/$0.36 \\
SuMa++°$^\ast$~\cite{SUMA++} & 0.64$/$0.22 & 1.60$/$0.46 & 1.00$/$0.37 & 0.67$/$0.46 & 0.37$/$0.26 & 0.40$/$0.20 & 0.46$/$0.21 & 0.34$/$0.19 & 1.10$/$0.35 & 0.47$/$0.23 & 0.66$/$0.28 & 0.70$/$0.29 \\
\midrule
KISS-ICP~\cite{KISS-ICP} & \underline{0.51}$/$\textbf{0.19} & 0.72$/$\underline{0.11} & 0.52$/$\textbf{0.15} & 0.66$/$\textbf{0.16} & \underline{0.35}$/$0.14 & 0.30$/$\underline{0.14} & \textbf{0.26}$/$\textbf{0.08} & 0.32$/$0.16 & 0.82$/$\textbf{0.18} & 0.49$/$\underline{0.13} & \underline{0.54}$/$0.19 & 0.50$/$\underline{0.15} \\ 
GenZ-ICP~\cite{GenZ-ICP} & 0.56$/$\phantom{0.00} & 0.80$/$\phantom{0.00} & \textbf{0.50}$/$\phantom{0.00} & 0.68$/$\phantom{0.00} & 0.40$/$\phantom{0.00} & \underline{0.28}$/$\phantom{0.00} & 0.28$/$\phantom{0.00} & 0.35$/$\phantom{0.00} & 0.81$/$\phantom{0.00} & 0.49$/$\phantom{0.00} & \textbf{0.50}$/$\phantom{0.00} & 0.51$/$\phantom{0.00}\\ 
LOAM~\cite{LOAM} & 0.78$/$\phantom{0.00} & 1.43$/$\phantom{0.00} & 0.92$/$\phantom{0.00} & 0.86$/$\phantom{0.00} & 0.71$/$\phantom{0.00} & 0.57$/$\phantom{0.00} & 0.65$/$\phantom{0.00} & 0.63$/$\phantom{0.00} & 1.12$/$\phantom{0.00} & 0.77$/$\phantom{0.00} & 0.79$/$\phantom{0.00} & 0.84$/$\phantom{0.00}\\ 
SA-LOAM$^\ast$~\cite{SA-LOAM} & 0.59$/$0.25 & 1.89$/$0.48 & 0.77$/$0.28 & 0.87$/$0.46 & 0.59$/$0.35 & 0.45$/$0.24 & 0.52$/$0.25 & 0.41$/$0.22 & 0.85$/$0.27 & 0.68$/$0.28 & 0.78$/$0.35 & 0.76$/$0.31 \\
SAGE-ICP$^\ast$~\cite{SAGE-ICP} & \underline{0.51}$/$\textbf{0.19} & \textbf{0.67}$/$\underline{0.11} & \underline{0.51}$/$0.16 & \underline{0.64}$/$\underline{0.17} & \textbf{0.33}$/$\underline{0.13} & 0.29$/$\underline{0.14} & \textbf{0.26}$/$0.09 & 0.32$/$\textbf{0.15} & 0.81$/$\textbf{0.18} & 0.47$/$0.14 & 0.55$/$\textbf{0.17} & \textbf{0.49}$/$\underline{0.15} \\
Ours (w$/$o sem.) & \underline{0.51}$/$0.20 & 0.86$/$0.12 & 0.55$/$0.16 & \textbf{0.56}$/$0.21 & 0.39$/$\textbf{0.10} & 0.29$/$0.15 & 0.27$/$\textbf{0.08} & \textbf{0.30}$/$\textbf{0.15} & \textbf{0.79}$/$0.21 & \textbf{0.42}$/$\underline{0.13} & 0.62$/$0.19 & 0.51$/$\underline{0.15} \\
Ours$^\ast$ & \textbf{0.50}$/$0.20 & \underline{0.71}$/$\textbf{0.09} & \underline{0.51}$/$\textbf{0.15} & 0.65$/$0.19 & 0.37$/$\textbf{0.10} & \textbf{0.26}$/$\textbf{0.13} & 0.27$/$0.09 & \textbf{0.30}$/$0.16 & \underline{0.80}$/$0.20 & \underline{0.45}$/$\textbf{0.12} & 0.56$/$\textbf{0.17} & \textbf{0.49}$/$\textbf{0.14} \\
\bottomrule
\end{tabular}
}
\end{table*}

Table~\ref{tab:results_kitti} shows the results on the KITTI Odometry Benchmark~\cite{KITTI}, supporting our first two claims. Even without semantic assistance, SOCC-ICP performs on par with state-of-the-art LiDAR odometry methods, achieving an RTE of 0.51\%. Incorporating semantic information from LSK3DNet~\cite{LSK3DNet} into mapping and registration improves translational performance on 6 of the 11 sequences, with minor degradation on 3, resulting in an overall RTE of 0.49\%. This indicates that semantic information provides a modest yet consistent benefit to SOCC-ICP's odometry accuracy.

\subsection{Comparison in General Environments}

\begin{table}[!t]
\centering
\caption{Quantitative results on the MulRan dataset~\cite{mulran}. We report the RTE in \%.}
\label{tab:results_mulran}

{\footnotesize 
\renewcommand{\arraystretch}{1.0} 
\begin{tabular}{C{1.9cm}|C{1.15cm}|C{1.15cm}|C{1.15cm}|C{1.15cm}}
\toprule
\textbf{Method} & \textbf{KAIST°} & \textbf{DCC°} & \textbf{Riverside°} & \textbf{Sejong°} \\ 
\midrule
MULLS°~\cite{MULLS} & 2.94 & 2.96 & 5.42 & 5.93 \\
\midrule
KISS-ICP~\cite{KISS-ICP} & 2.28 & \underline{2.34} & \underline{2.89} & 4.69 \\
GenZ-ICP~\cite{GenZ-ICP} & \underline{2.27} & 2.39 & 3.01 & \underline{4.62} \\
Ours (w/o sem.) & \textbf{2.17} & \textbf{2.11} & \textbf{2.74} & \textbf{4.25} \\
\bottomrule
\end{tabular}
}
\end{table}

\begin{table}[!t]
\centering
\caption{Quantitative results on the Newer College dataset~\cite{NewerCollege}. We report the RTE in \%.}
\label{tab:res_newer_college}

{
\renewcommand{\arraystretch}{1.0} 
\begin{tabular}{C{2.5cm}|C{2.4cm}|C{2.4cm}}
\toprule
\textbf{Method} & \textbf{short experiment°} & \textbf{long experiment°} \\ 
\midrule
MULLS°~\cite{MULLS} & 0.82 & 1.23 \\
CT-ICP°~\cite{CT-ICP} & 0.48 & 0.58 \\
\midrule
KISS-ICP~\cite{KISS-ICP} & 0.51 & 0.96 \\
GenZ-ICP~\cite{GenZ-ICP} & \underline{0.46} & \textbf{0.94} \\
Ours (w/o sem.) & \textbf{0.45} & \textbf{0.94} \\
\bottomrule
\end{tabular}
}
\end{table}

Each MulRan sequence~\cite{mulran} was recorded three separate times. In Table~\ref{tab:results_mulran}, we report the average RTE per sequence. SOCC-ICP consistently outperforms all other methods by a clear margin. As shown in Table~\ref{tab:res_newer_college}, SOCC-ICP also achieves the best performance on the Newer College short experiment~\cite{NewerCollege}. On the long experiment, SOCC-ICP matches GenZ-ICP as the best-performing LiDAR odometry method, while the SLAM system CT-ICP attains the lowest error overall, likely due to global pose corrections enabled by loop closure~\cite{GenZ-ICP}. Overall, these results support our first claim and demonstrate that SOCC-ICP delivers reliable odometry even without semantic assistance.

\subsection{Comparison in Corridor Scenarios}

\begin{table}[!t]
\centering
\caption{Quantitative results for the Corridor1 and Corridor2 sequences of the Ground-Challenge dataset~\cite{GroundChallenge}, and the Long Corridor sequence of the SubT-MRS dataset~\cite{SubT-MRS}. We report APE and RPE for the translational component.}
\label{tab:res_ground_chall_subt_mrs}

{\scriptsize 
\setlength{\tabcolsep}{2pt} 
\renewcommand{\arraystretch}{1.0} 
\begin{tabular}{C{1.19cm}|C{1.65cm}|C{0.54cm}C{0.54cm}C{0.54cm}C{0.54cm}|C{0.54cm}C{0.54cm}C{0.54cm}C{0.54cm}}
\toprule
\textbf{Dataset} & & \multicolumn{4}{c|}{\textbf{Abs. Pose Error [m]}} & \multicolumn{4}{c}{\textbf{Rel. Pose Error [m]}} \\
\textbf{Sequence} & \textbf{Method} & \textbf{Mean} & \textbf{Max} & \textbf{RMSE} & $\bm{\sigma}$ & \textbf{Mean} & \textbf{Max} & \textbf{RMSE} & $\bm{\sigma}$ \\ 
\midrule
\textbf{Grd-Chall} & CT-ICP~\cite{CT-ICP} & 0.44 & 1.05 & 0.54 & 0.30 & 0.05 & \textbf{0.23} & \underline{0.06} & \underline{0.04}\\ 
Corridor1 & KISS-ICP~\cite{KISS-ICP} & 1.70 & 4.76 & 2.17 & 1.35 & 0.12 & 0.59 & 0.15 & 0.09 \\
(zigzag) & GenZ-ICP~\cite{GenZ-ICP} & \underline{0.19} & \underline{0.49} & \underline{0.24} & \underline{0.14} & \underline{0.04} & \textbf{0.23} & \underline{0.06} & \underline{0.04} \\
\cmidrule{2-10}
 & Ours (w/o sem.) & \textbf{0.08} & \textbf{0.22} & \textbf{0.09} & \textbf{0.03} & \textbf{0.02} & \textbf{0.23} & \textbf{0.04} & \textbf{0.03} \\
\midrule
\midrule
\textbf{Grd-Chall} & CT-ICP~\cite{CT-ICP} & 1.04 & 2.36 & 1.30 & 0.78 & \underline{0.12} & \underline{0.35} & \underline{0.14} & \underline{0.06}\\ 
Corridor2 & KISS-ICP~\cite{KISS-ICP} & 0.54 & 1.34 & 0.68 & 0.41 & 0.14 & 0.46 & 0.16 & 0.08 \\
(straight) & GenZ-ICP~\cite{GenZ-ICP} & \underline{0.18} & \underline{0.41} & \underline{0.20} & \underline{0.09} & \underline{0.12} & 0.36 & \underline{0.14} & 0.07 \\
\cmidrule{2-10}
& Ours (w/o sem.) & \textbf{0.07} & \textbf{0.21} & \textbf{0.08} & \textbf{0.04} & \textbf{0.02} & \textbf{0.18} & \textbf{0.03} & \textbf{0.02} \\
\midrule
\midrule
\textbf{SubT-MRS} & CT-ICP~\cite{CT-ICP} & 44.18 & 60.14 & 45.66 & 11.55 & 0.19 & 7.15 & 0.68 & 0.65 \\ 
Long Corr. & KISS-ICP~\cite{KISS-ICP} & 6.83 & 19.05 & 8.72 & 5.41 & 0.10 & 0.94 & 0.14 & 0.10 \\ 
 & GenZ-ICP~\cite{GenZ-ICP} & \textbf{1.69} & \textbf{4.32} & \textbf{1.99} & \underline{1.04} & \textbf{0.06} & \underline{0.73} & \textbf{0.09} & \textbf{0.07} \\
\cmidrule{2-10}
 & Ours (w/o sem.) & \underline{1.72} & \underline{4.41} & \underline{2.00} & \textbf{1.01} & \underline{0.07} & \textbf{0.64} & \underline{0.10} & \textbf{0.07} \\
\bottomrule

\end{tabular}
}
\end{table}

Lee et al.~\cite{GenZ-ICP} used corridor sequences from the Ground-Challenge~\cite{GroundChallenge} and SubT-MRS~\cite{SubT-MRS} datasets to demonstrate the robustness of GenZ-ICP in degenerate environments, where single-metric methods tend to degrade. This behavior is evident in Table~\ref{tab:res_ground_chall_subt_mrs}, where CT-ICP and KISS-ICP generally underperform compared to GenZ-ICP and SOCC-ICP. On the Ground-Challenge sequences, SOCC-ICP achieves the best absolute and relative pose errors, outperforming GenZ-ICP, while on the SubT-MRS Long Corridor sequence it attains comparable accuracy. These results support our third claim and show that, by adopting the adaptive combination of point-to-point and point-to-plane residuals of GenZ-ICP, SOCC-ICP inherits similar robustness.

\subsection{Ablation Study}

We evaluate individual components of SOCC-ICP on KITTI, reporting average RTE [\%] and RRE [deg/100\,m]. 
Recall that full SOCC-ICP achieves 0.491/0.144 (see Table~\ref{tab:results_kitti}).
\textbf{Voxel anchor point.}
We evaluate the choice of anchor point representing each map voxel in the ICP correspondences.
Replacing the first-inserted point $\mathbf{q}^{(1)}_i$ with the voxel center degrades performance drastically (0.985/0.480), confirming that retaining point-level information within each voxel is essential. 
Using the running mean point instead also degrades performance (0.508/0.151), as it shifts with newly integrated measurements unlike the more stable first-inserted point.
Combining the mean point with exponential decay of the voxel's mean and covariance statistics to emphasize recent observations yields 0.494/0.148.
\textbf{Cleaning ray.}
Disabling raycasting-based free-space updates increases the RTE from 0.491 to 0.501 with an RRE of 0.145, confirming that dynamic object removal via the occupancy grid benefits accuracy.
\textbf{Weighting and downsampling.}
Removing the occupancy-based weight $w_{\text{occ}}$ yields 0.494/0.144, and removing the semantic consistency weight $w_{\text{sem}}$ yields 0.495/0.142.
Downsampling without using semantic information results in 0.498/0.149.
Each component contributes a modest improvement, and their combination in the full system achieves the best overall performance.

\subsection{Qualitative Mapping Results}

\begin{figure}[t]
    \centering
    \hfill
    \begin{subfigure}[t]{0.445\columnwidth}
        \centering
        \includegraphics[
            width=\linewidth,
            trim=0mm 3mm 0mm 3mm,
            clip
        ]{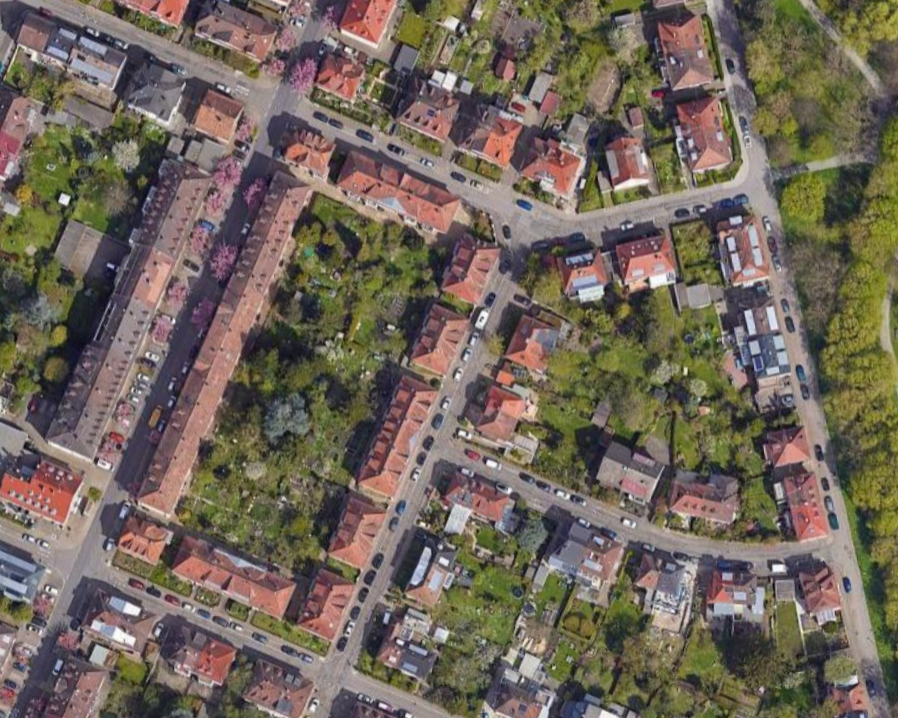}
        \caption{Satellite image containing the trajectory (KITTI 07).}
    \end{subfigure}
    \hfill
    \hfill
    \begin{subfigure}[t]{0.485\columnwidth}
        \centering
        \includegraphics[
            width=\linewidth,
            trim=0mm 0mm 0mm 0mm,
            clip
        ]{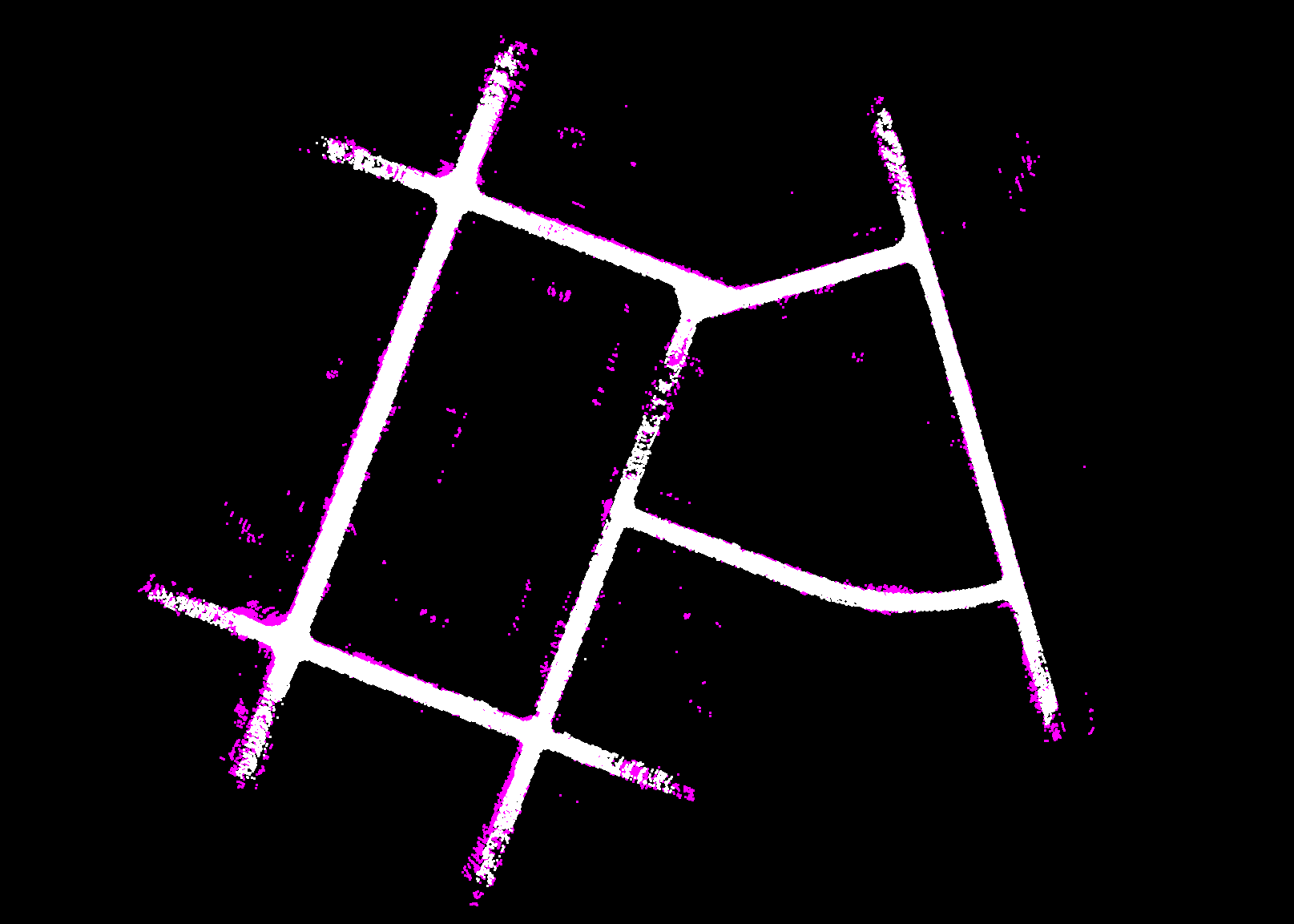}
        \caption{GT (white, foreground) and SOCC-ICP (pink) road voxels.}
    \end{subfigure}
    
    \vspace{0.5em}
 
    \begin{subfigure}[t]{0.485\columnwidth}
        \centering
        \includegraphics[
            width=\linewidth,
            trim=0mm 0mm 0mm 0mm,
            clip
        ]{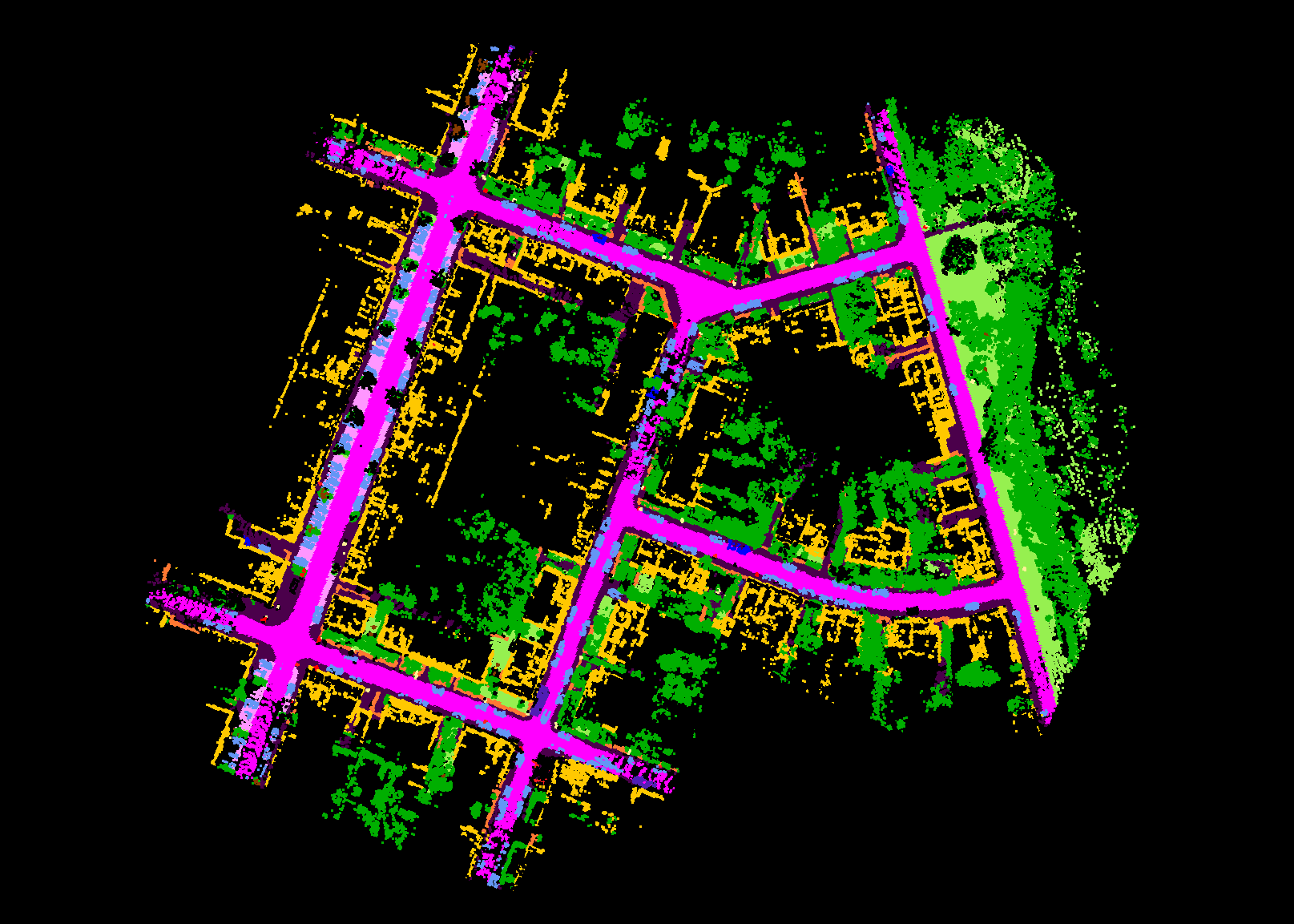}
        \caption{GT mapping result.}
    \end{subfigure}
     \hfill
    \begin{subfigure}[t]{0.485\columnwidth}
        \centering
        \includegraphics[
            width=\linewidth,
            trim=0mm 0mm 0mm 0mm,
            clip
        ]{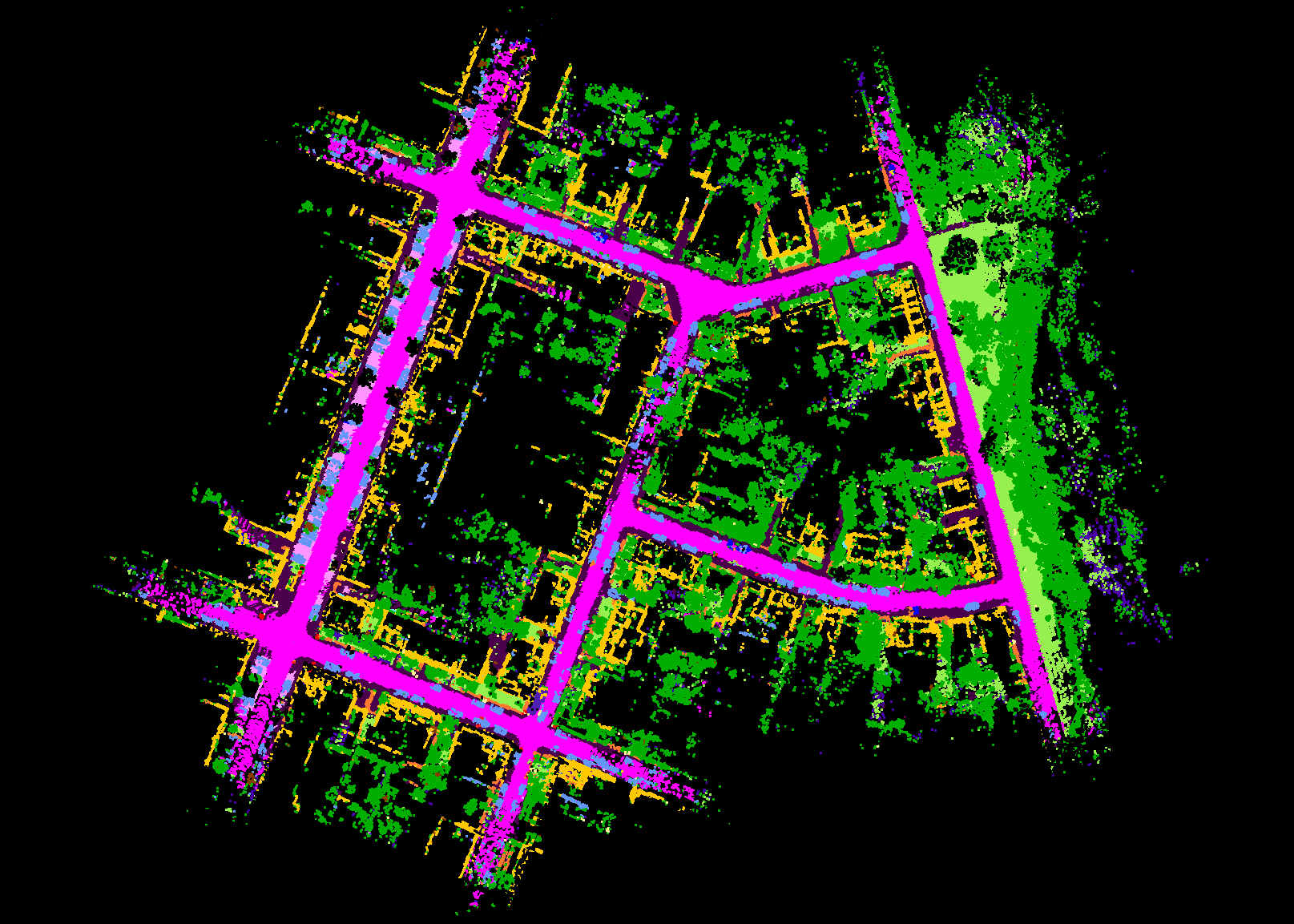}
        \caption{SOCC-ICP mapping result.}
    \end{subfigure}

    \caption{Comparison of occupancy grid mapping results.}
    \label{fig:mapping_result}
\end{figure}

For additional qualitative evaluation, Figure~\ref{fig:mapping_result}~(d) shows the mapping results on KITTI sequence~07 obtained with SOCC-ICP. Despite operating without loop closure, the semantic occupancy grid constructed from predicted semantics and estimated odometry closely matches the reference grid built from ground-truth KITTI odometry and SemanticKITTI annotations, indicating low drift over long trajectories.

\section{Conclusion}
We presented SOCC-ICP, the first LiDAR odometry framework to perform online scan-to-map registration on a 3D semantic occupancy grid. 
Our experiments across urban, campus, and corridor environments confirm that an occupancy grid suffices as a sole map representation for competitive LiDAR odometry, with adaptive residual selection and optional semantic cues providing complementary gains in robustness and accuracy.
Beyond odometry, this work demonstrates that scan alignment and semantic occupancy grid mapping can be performed jointly within a single representation, eliminating redundant map structures, naturally handling dynamic objects, and forming an integrated system that directly supports downstream robotic applications.  
Future work includes improving computational efficiency and extending the approach toward LiDAR-inertial odometry or a complete SLAM system.